\documentclass[10pt]{article} 
\usepackage[preprint]{tmlr}

\usepackage{amsmath,amsfonts,bm}

\def\eqref#1{equation~\ref{#1}}

\def\1{\bm{1}}

\DeclareMathAlphabet{\mathsfit}{\encodingdefault}{\sfdefault}{m}{sl}
\SetMathAlphabet{\mathsfit}{bold}{\encodingdefault}{\sfdefault}{bx}{n}

\usepackage[T1]{fontenc}
\usepackage[utf8]{inputenc}
\usepackage{microtype}
\usepackage{inconsolata}
\usepackage{times}

\usepackage{graphicx}
\usepackage{float}
\usepackage{booktabs}
\usepackage{multirow}
\usepackage{enumitem}
\usepackage{amsmath,amssymb}
\usepackage{afterpage}
\usepackage{diagbox}
\usepackage{subcaption} 
\usepackage{wrapfig}
\usepackage{url}
\usepackage{verbatim}
\usepackage[most]{tcolorbox} 
\usepackage{colortbl}
\usepackage{ulem}
\usepackage{fancyhdr}
\usepackage{hyperref}

\definecolor{graybg}{gray}{.95}
\definecolor{bgcolor}{RGB}{246,248,250}

\fancypagestyle{firstpage}{
    \fancyfoot[L]{\scriptsize
        \url{https://github.com/CedricPei/Feather-SQL} \\
        \textsuperscript{\textdagger} Corresponding author.
    }
}

\title{Feather-SQL: A Lightweight NL2SQL Framework with Dual-Model Collaboration Paradigm for Small Language Models}

\author{
\textbf{\fontsize{12pt}{14pt}\selectfont
Wenqi Pei$^1$, 
Hailing Xu$^1$, 
Hengyuan Zhao$^1$, 
Shizheng Hou$^1$,
Han Chen$^1$,
}
\\[6pt]
\textbf{\fontsize{12pt}{14pt}\selectfont
Zining Zhang$^1$, 
Pingyi Luo$^2$, 
and Bingsheng He $^1$\textsuperscript{,\textdagger}}
\\[6pt]
\textsuperscript{1} {\fontsize{12pt}{14pt}\selectfont National University of Singapore},
\textsuperscript{2} {\fontsize{12pt}{14pt}\selectfont 4Paradigm}
}

\def\month{MM}  
\def\year{YYYY} 
\def\openreview{\url{https://openreview.net/forum?id=XXXX}} 

\begin{document}

\maketitle

\begin{abstract}
Natural Language to SQL (NL2SQL) has seen significant advancements with large language models (LLMs). However, these models often depend on closed-source systems and high computational resources, posing challenges in data privacy and deployment. In contrast, small language models (SLMs) struggle with NL2SQL tasks, exhibiting poor performance and incompatibility with existing frameworks. To address these issues, we introduce \textbf{Feather-SQL}, a new lightweight framework tailored for SLMs. Feather-SQL improves SQL executability and accuracy through 1) schema pruning and linking, 2) multi-path and multi-candidate generation. Additionally, we introduce the \textbf{1+1 Model Collaboration Paradigm}, which pairs a strong general-purpose chat model with a fine-tuned SQL specialist, combining strong analytical reasoning with high-precision SQL generation. Experimental results on BIRD demonstrate that Feather-SQL improves NL2SQL performance on SLMs, with around 10\% boost for models without fine-tuning. The proposed paradigm raises the accuracy ceiling of SLMs to 54.76\%, highlighting its effectiveness.
\end{abstract}

\thispagestyle{firstpage}

\section{Introduction}
\label{introduction}
\begin{figure}[htbp]
\centering
\includegraphics[width=0.8\linewidth]{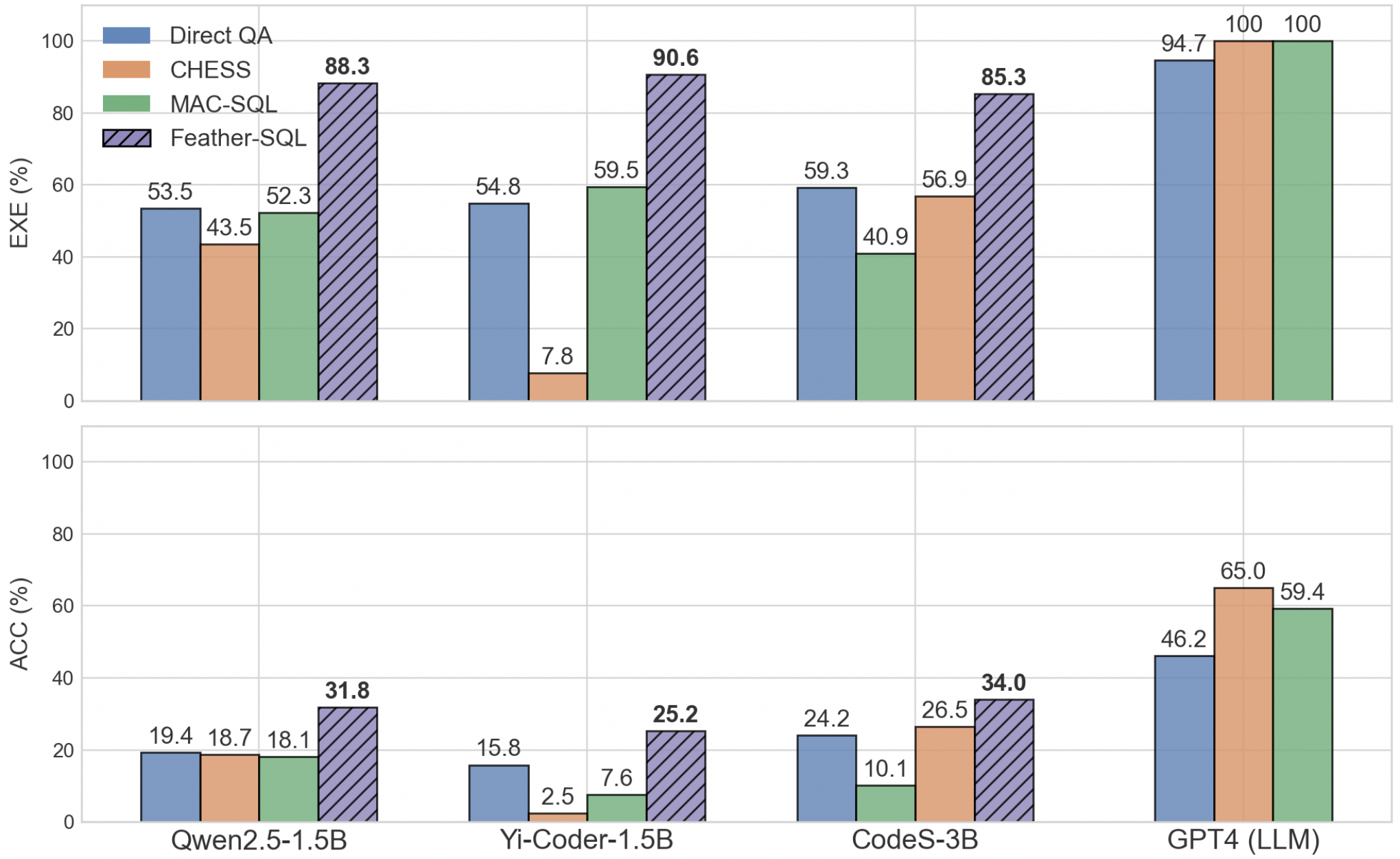}
\caption{NL2SQL performance on the BIRD DEV dataset. \underline{EXE} (Executability) measures successful query executions, while \underline{ACC} (Accuracy) measures correct result matches.}
\label{fig:acc}
\end{figure}

Natural Language to SQL (NL2SQL) is the task of converting natural language questions into corresponding SQL queries, allowing users to retrieve structured data from databases without requiring proficiency in SQL language. In recent years, the field has seen significant advancements with the emergence of large language models (LLMs) such as GPT-4 \citep{gpt4}, enabling frameworks like CHASE-SQL \citep{chase-sql} and XiYan-SQL \citep{xiyan} to achieve state-of-the-art (SOTA) performance. However, two limitations hinder their practical adoption. First, mainstream methods depend on closed-source models, and their reliance on external APIs introduces data privacy risks in sensitive domains like healthcare and finance \citep{survey1}. Second, most open-source research focuses on models with 7B--30B parameters, leaving small language models (SLMs) with 4B or fewer parameters relatively underexplored. Meanwhile, many relational databases are deployed on high-performance systems with limited GPU resources. With efficient inference frameworks (e.g., Intel IPEX-LLM \citep{ipex}) or quantization techniques, SLMs can help drive the broader adoption of NL2SQL in real-world scenarios while preserving data privacy. 

\begin{wrapfigure}{l}{0.5\textwidth}
\includegraphics[width=\linewidth]{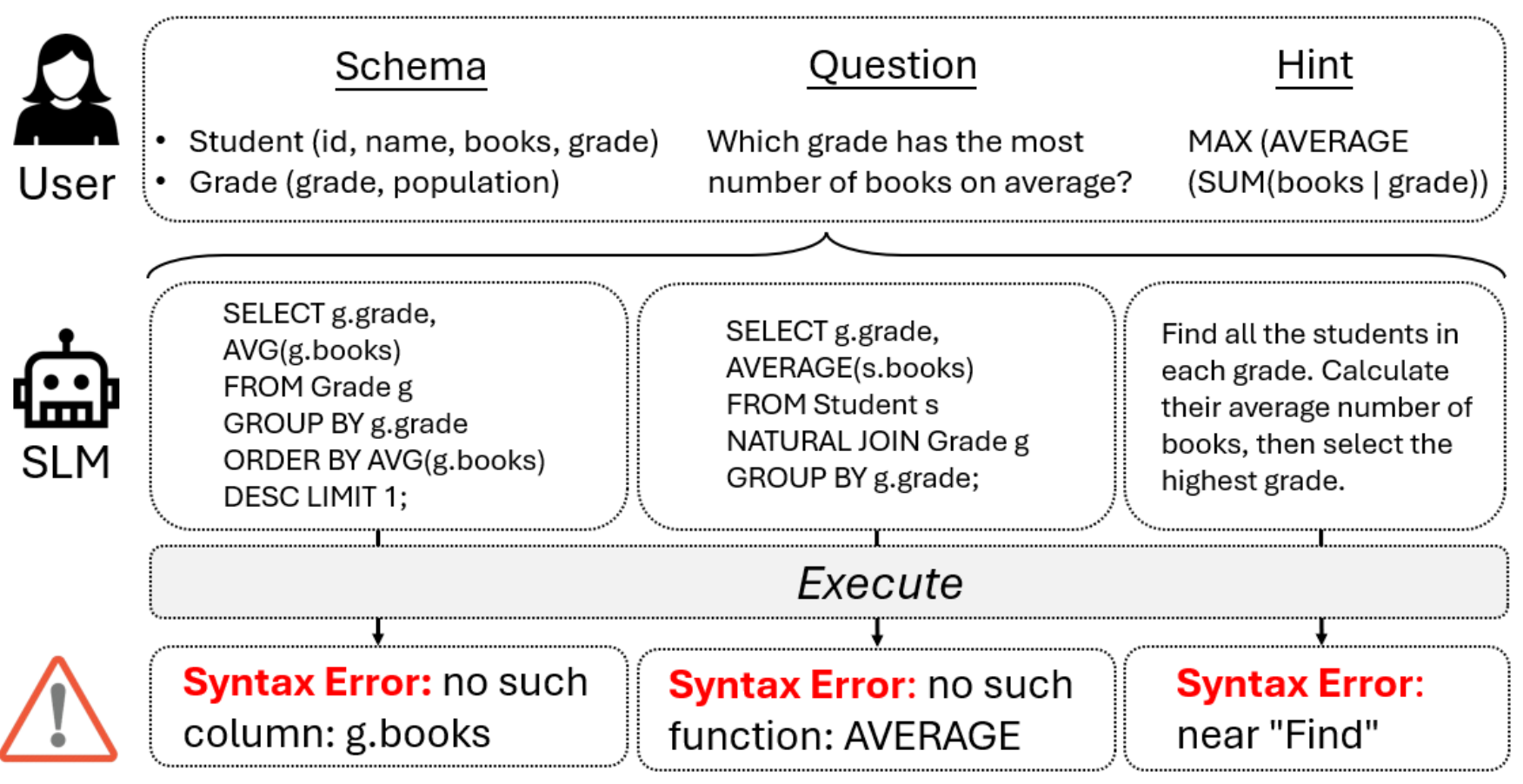}
\caption{Examples of typical syntax errors produced by small language models (SLMs) in an NL2SQL scenario.}
\label{fig:error}
\end{wrapfigure}

\begin{wrapfigure}{r}{0.5\textwidth}
\includegraphics[width=\linewidth]{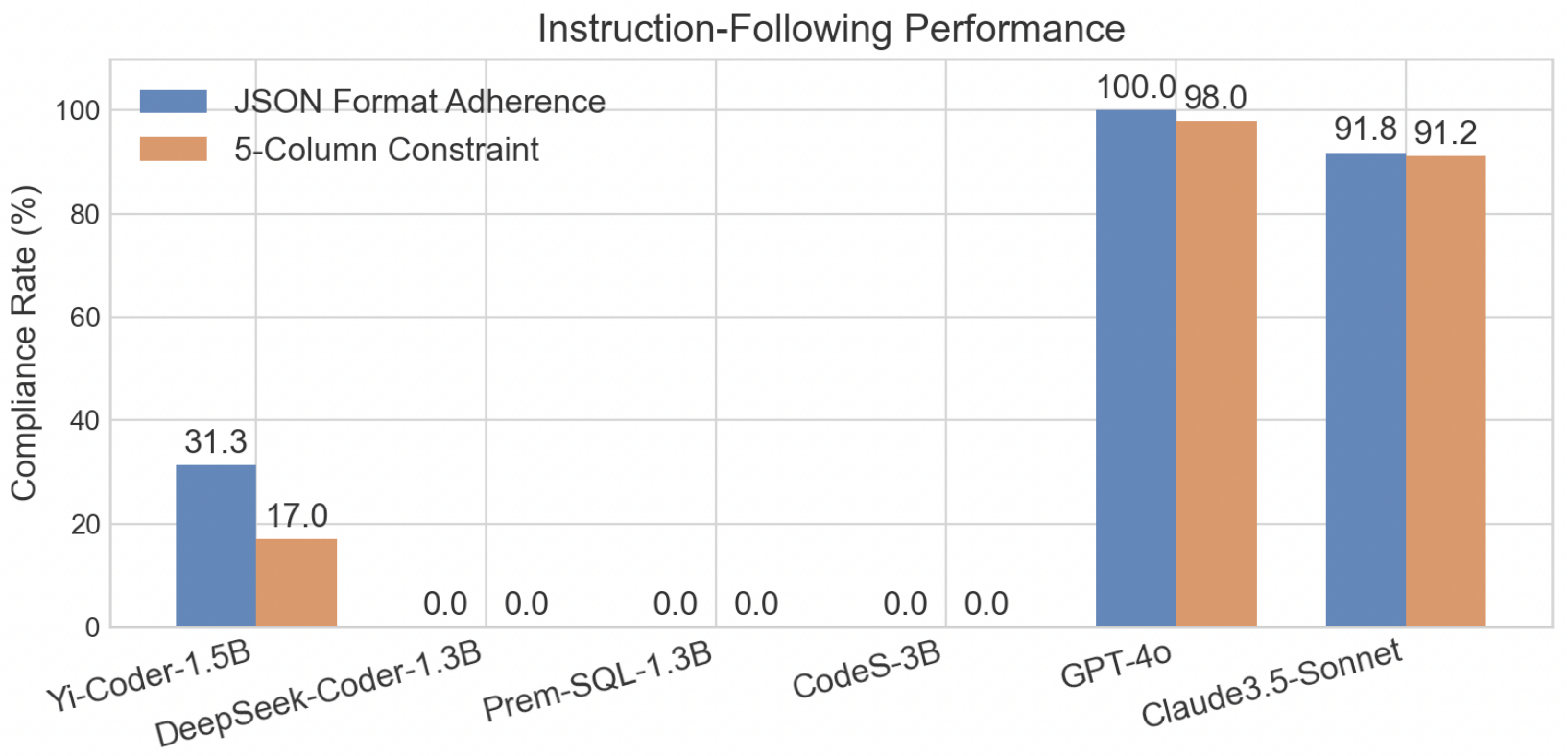}
\caption{Experiments conducted on a CHESS-provided BIRD subset for schema linking. Models are required to output a JSON-formatted response containing no more than five relevant columns related to the question, without generating any extraneous content.}
\label{fig:instruction}
\vspace{-5mm}
\end{wrapfigure}

In this paper, we focus on enhancing NL2SQL performance using SLMs. As shown in Figure \ref{fig:acc}, SLMs face two key challenges: (1) \textbf{one critical issue is their sharp decline in executability}. Unlike LLMs, which can effectively handle long-context dependencies, SLMs struggle with complex database schema and verbose prompts, often leading to confusion or hallucinated outputs \citep{ slm_survey, tasql} (Figure \ref{fig:error}); (2) \textbf{existing frameworks for NL2SQL tasks with LLMs are incompatible with SLMs}, as they rely on strong instruction-following capabilities to produce intermediate results, which SLMs lack. As illustrated in Figure \ref{fig:instruction}, SLM outputs frequently violate imposed requirements: they often fail to conform to JSON or array specifications and do not meet predefined constraints. Directly applying these frameworks to SLMs may further degrade executability.

To address these challenges, we propose \textbf{Feather-SQL}, a lightweight framework tailored for SLMs to enhance both executability and accuracy in NL2SQL tasks. Feather-SQL consists of six key components: schema pruning, schema linking, multi-path generation, multi-candidate generation, correction, and selection. Designed specifically for SLMs, schema pruning streamlines input processing by discarding irrelevant tables, allowing models to concentrate on essential database elements. Schema linking improves alignment between questions and database schema, ensuring accurate column selection. To mitigate errors from linking and pruning, multi-path generation explores diverse query formulation strategies, enhancing robustness. Multi-candidate generation further improves the model's executability and accuracy by enhancing the variety of generated SQL queries, thereby increasing the likelihood of producing correct candidates. The best candidate is then selected through execution validation and ranking. Complementing these components, we introduce extraction and simplification prompting strategies. Extraction selectively retrieves key information, while simplification removes extraneous prompt details to lower computational overhead. By integrating these techniques, Feather-SQL enables SLMs to generate SQL queries more reliably despite their inherent limitations.

A common approach to enhancing SLMs is fine-tuning. However, while fine-tuned SLMs for SQL generation tasks (e.g., Prem-SQL \citep{premsql}, CodeS \citep{codes}) outperform general-purpose chat models on core NL2SQL tasks, they suffer from catastrophic forgetting \citep{luo2025empiricalstudycatastrophicforgetting,kotha2024understanding} on auxiliary tasks—where task-specific fine-tuning erodes their foundational reasoning abilities. To counter this, we propose \textbf{1+1 Model Collaboration Paradigm}, in which a general-purpose chat model handles reasoning-intensive auxiliary tasks (e.g., schema linking and candidate selection), while a fine-tuned SQL specialist focuses on query generation. This collaboration leverages both models' strengths: the general model provides broad reasoning
ability, while the specialist delivers domain-specific precision. Experiments confirm that the paradigm improves overall performance. Our main contributions are as follows:

\begin{itemize} 
\item We introduce Feather-SQL, an NL2SQL framework for SLMs to address their unique challenges of low executability and incompatibility with existing LLM-based frameworks. 

\item We propose a novel 1+1 Model Collaboration paradigm that mitigates catastrophic forgetting in fine-tuned SLMs by delegating reasoning-intensive tasks to a general-purpose chat model.

\item Extensive experiments on the Spider and BIRD datasets demonstrate that Feather-SQL consistently achieves strong performance with various SLMs, and when paired with the paradigm, it yields SOTA results on BIRD within the scope of SLMs. 
\end{itemize}

\section{Related Work}

\subsection{Conventional Methods}
Early NL2SQL systems were rule- or template-based \citep{chill,nalir,athena}.  
Although effective on small, curated datasets, these approaches demanded extensive manual engineering and did not generalise well.  
The arrival of sequence-to-sequence (Seq2Seq) neural models marked a shift to data-driven methods. Models such as Seq2SQL \citep{wikisql}, SQLNet \citep{xu2017sqlnet}, IRNet \citep{jha2019irnet}, RyanSQL \citep{choi2021ryansql}, and RESDSQL \citep{li2023resdsql} jointly encode the natural-language question and database schema before decoding the corresponding SQL query.  Fine-tuning pretrained language models (PLMs)—for example, BERT \citep{bert} and T5 \citep{T5}, as used in Graphix-T5 \citep{li2023graphix}—further improves robustness, yet still requires substantial annotated data and struggles with highly complex schemas.

\subsection{LLM and SLM Approaches}
Instruction-tuned large language models (LLMs) now achieve state-of-the-art performance by decomposing NL2SQL into subtasks.  
Systems such as DIN-SQL \citep{dinsql}, TASQL \citep{tasql}, MAC-SQL \citep{macsql}, and CHESS \citep{chess} exceed earlier accuracy, but their multi-stage prompting incurs significant computation, and potential privacy risks when queries leave the user’s environment.

To alleviate these drawbacks, researchers have turned to small language models (SLMs). Approaches such as CodeS \citep{codes}, DTS-SQL \citep{sl6}, Prem-SQL \citep{premsql}, and SQLCoder \citep{sqlcoder} fine-tune SLMs on NL-to-SQL datasets. However, training makes them susceptible to catastrophic forgetting, diminishing their compositional-reasoning ability. MSc-SQL \citep{mscsql} trains separate $\sim$10B-parameter models for different subtasks to preserve capabilities, but at the expense of extra memory and storage, limiting practical deployment. Therefore, a lightweight framework that empowers SLMs to perform NL2SQL effectively—without prohibitive resource demands—remains an open and important research goal.

\section{Methodology}
\subsection{Feather-SQL}

\begin{figure*}[htbp]
\centering
\includegraphics[width=\linewidth]{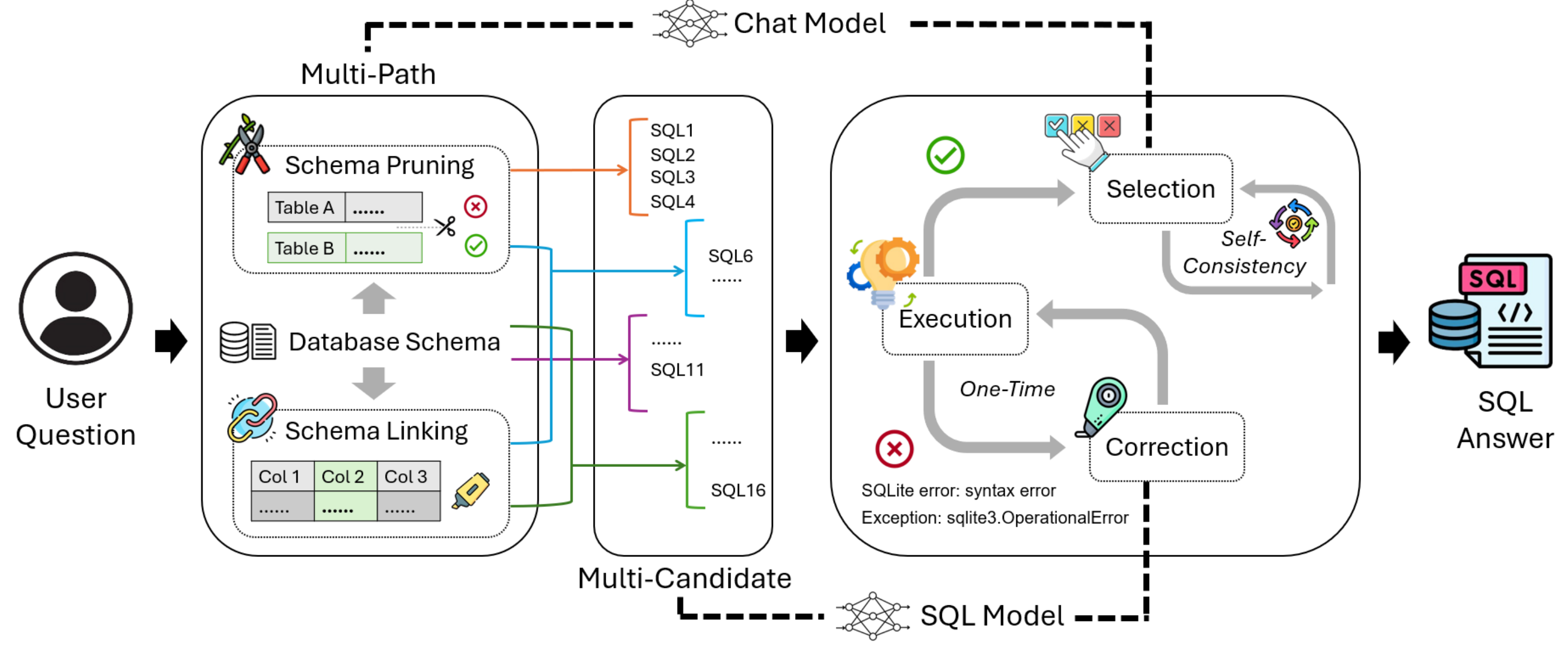}
\caption{An overview of the Feather-SQL framework for small language models (SLMs) in NL2SQL tasks. The pipeline comprises six core modules---\textit{schema pruning}, \textit{schema linking}, \textit{multi-path generation}, \textit{multi-candidate generation}, \textit{correction}, and \textit{selection}---which collaboratively boost query executability and accuracy. Additionally, the \textit{1+1 Model Collaboration Paradigm} pairs a general-purpose Chat Model with a SQL fine-tuned Model: the Chat Model conducts the Multi-path and Selection stages (upper dashed links), while the SQL Model performs the Multi-Candidate and Correction stages (lower dashed links).}
\label{fig:framework}
\end{figure*}

As shown in Figure \ref{fig:framework}, we propose Feather-SQL to enhance the performance of SLMs in NL2SQL. This framework comprises several components, including Schema Pruning, Multi-Path, and Multi-Candidate Generation, which are specifically designed to address the challenges of SLMs. We will elaborate on these components in the following sections.

\noindent\textbf{Schema Pruning.} This step dynamically reduces schema complexity by identifying and filtering out tables semantically irrelevant to the user’s question. Only the Data Definition Language (DDL) statements of tables judged pertinent advance to later stages in the pipeline, preventing small language models from being overwhelmed by lengthy inputs while preserving essential information. Although pruning was previously explored by \cite{jose2023multilingualtranslatorsqldatabase}— who applied it as an offline, training-time preprocessing step driven by statistical analysis—our approach performs it on-the-fly at inference using one SLM.

\noindent\textbf{Schema Linking.} This step aligns the question with the database schema by identifying relevant columns through semantic analysis \citep{jha2019irnet}. As a commonly adopted practice, schema linking extracts pertinent columns from the complete schema based on the given question, facilitating downstream processing \citep{ratsql, chess}. By establishing precise mappings between natural language expressions and database elements, this process significantly enhances SQL generation accuracy.

\noindent\textbf{Multi-Path Generation.} This step employs four distinct prompt types: (1) with both schema linking and pruning, (2) linking only, (3) pruning only, and (4) without either operation. The multi-path design mitigates the risk of information loss caused by pruning errors and reduces potential misunderstandings arising from linking inaccuracies.

\noindent\textbf{Multi-Candidate Generation.} This step generates multiple SQL queries in parallel to increase the likelihood of producing a correct result \citep{chase-sql, mscsql}. To ensure diversity, beam search is employed alongside carefully tuned temperature and top-p parameters. Each path consistently generates a fixed number of candidate queries, maintaining a balanced exploration of possible solutions.

\begin{figure}[H]
  \centering
  \includegraphics[width=\linewidth]{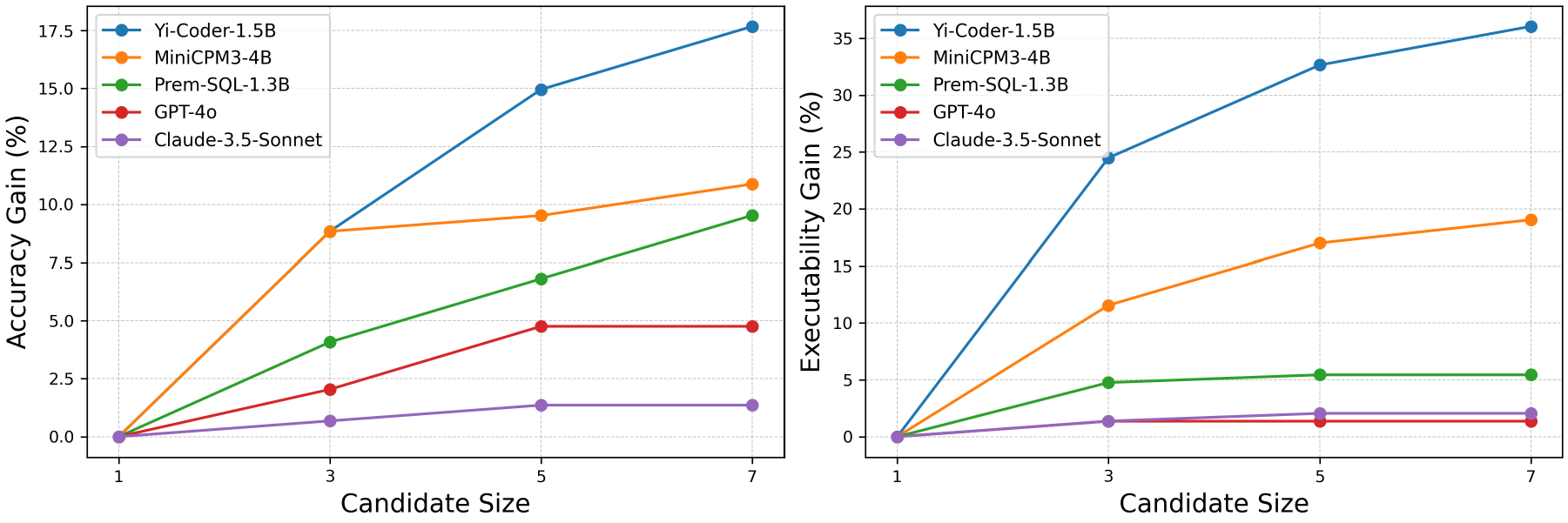}
  \caption{
  Accuracy gain and executability gain by candidate size. Gains are measured as the percentage-point difference from each model’s performance with a single candidate. 
  For both metrics, a set of candidates is counted as correct or executable if at least one candidate in the set meets the criterion.
  }
  \label{fig:gains}
\end{figure}

As shown in Figure~\ref{fig:gains}, increasing candidate size yields consistent improvements in both accuracy and executability for SLMs, with notably larger gains compared to LLMs. Larger models are already robust with a single candidate and show only marginal improvements when more candidates are provided. (Details in Appendix~\ref{app:mutli-candidate}.)

\noindent\textbf{Correction.} This step executes each generated query and handles it based on the outcome \citep{macsql, dinsql}. If a query executes successfully, it is directly added to the array of executable SQL queries. For failed queries, error feedback is used to revise the query through a self-correction approach, generating two new candidate queries. If any of these revised queries are executable, they are also stored in the array of executable SQL queries.

\noindent\textbf{Selection.} This step applies a selection-ranking method to assess all executable queries according to their alignment with the expected answers \citep{chase-sql,xiyan, chess}. If a query yields a limited number of results, the evaluation considers both the query and its execution outcome. In contrast, the evaluation focuses solely on the query itself. The selection process is repeated three times, and the mode of the rankings is returned as the final result.

\subsection{Prompting Strategies}
\noindent\textbf{Extraction.} As mentioned in Section \ref{introduction}, SLMs struggle to meet structural constraints, thus we propose an extraction strategy to avoid rigid structural outputs by allowing SLMs to freely generate responses. 
This improves accuracy on reasoning tasks by bypassing syntactic constraints. We have two methods to achieve that: \textbf{(1) Lexical Matching:} This method identifies valid schema elements by matching table/column names explicitly mentioned in the natural language response against the database schema. For instance, when the SLM outputs "The required tables include customer and orders", the system verifies and extracts customer/orders only if they exist in the schema. \textbf{(2) Pattern Matching:} This method extracts the final answer by identifying predefined patterns in the model’s output, such as "answer is" or "Answer:". 
\begin{wraptable}{r}{0.4\textwidth}
  \centering
  \renewcommand{\arraystretch}{1}
  \resizebox{\linewidth}{!}{
  \begin{tabular}{llr}
    \toprule
    \textbf{Method} & \textbf{Stage} & \textbf{Words} \\
    \midrule
    \multirow{4}{*}{CHESS}
      & Information Retrieval & 423 \\
      & Schema Selection      & 2522 \\
      & Candidate Generation  & 4888 \\
      & Revision              & 1835 \\
    \midrule
    \multirow{3}{*}{MAC-SQL}
      & Selection   & 552 \\
      & Decomposition & 836 \\
      & Revision    & 174 \\
    \midrule
    \multirow{5}{*}{Feather-SQL}
      & Schema Pruning  & 267 \\
      & Schema Linking  & 287 \\
      & Generation      & 190 \\
      & Correction      & 106 \\
      & Selection       & 271 \\
    \bottomrule
  \end{tabular}
  }
  \caption{Prompt length comparison.}
  \label{tab:stages}
  \vspace{-1.5cm}
\end{wraptable}
For example, if the model generates “The answer is 128", the framework detects the pattern and extracts "128" as the final result.

\noindent\textbf{Simplification.} We reduce computational overhead by minimizing prompt verbosity while keeping the task unambiguous. In Feather-SQL, we achieve this by removing superfluous details and using concise instructions with the fewest effective examples. This approach refines the input by eliminating unnecessary complexity, avoiding the need for SLMs to process lengthy inputs while maintaining the clarity of the task. 

As shown in Table~\ref{tab:stages}, CHESS uses very long, instruction-heavy prompts, and MAC-SQL also exceeds 500 words in 2 stages. Only Feather-SQL stays concise across all stages, balancing context and complexity without burdening SLMs with lengthy inputs.

\subsection{1+1 Collaboration Paradigm}

Our paradigm categorizes NL2SQL pipeline tasks into two types: reasoning-intensive tasks and SQL generation tasks. Reasoning tasks, such as schema linking and candidate evaluation, require strong contextual understanding and adaptability, while SQL generation demands precision in query synthesis. To optimize performance, we employ two specialized models: the general-purpose chat Model for reasoning tasks and the SQL fine-tuned model for SQL generation. By leveraging their complementary strengths, our approach improves overall NL2SQL accuracy and robustness.  

\noindent\textbf{General-purpose Chat Model.}
This model is designed for reasoning-intensive tasks, leveraging broad linguistic and contextual comprehension without domain-specific fine-tuning, which helps prevent catastrophic forgetting. Compared to the SQL Specialist Model, it is more effective in schema linking and evaluating SQL candidates, ensuring that the SQL generation process is guided by accurate and well-structured contextual information. 

\noindent\textbf{SQL Fine-tuned Model.}  
Optimized exclusively for SQL generation, this model is extensively trained on large-scale NL2SQL datasets, allowing it to achieve superior performance on SQL-specific tasks. Its focused training reduces hallucinations and enhances both query executability and accuracy.

\section{Experiments}
\subsection{Settings}
\subsubsection{Datasets}
\noindent \textbf{BIRD} \citep{bird} as a representative and challenging benchmark dataset for NL2SQL, encompasses databases over 37 professional domains. Due to the proprietary nature of the BIRD TEST dataset, we conduct our experiments using the publicly accessible BIRD DEV subset, which contains 1,534 unique question-SQL pairs.

\noindent \textbf{Spider} \citep{spider} is another large-scale benchmark dataset for cross-domain SQL generation, covering 138 different domains. Compared to BIRD, Spider is relatively simpler, as its SQL structures and schema are generally less complex. Our experiments utilize the SPIDER TEST set, comprising 2,147 question-SQL pairs.

\subsubsection{Evaluation Metrics}
\noindent\textbf{Execution Accuracy (EX)} \citep{bird} is a widely adopted metric in NL2SQL evaluations, measuring whether the result of executing the generated query matches the result of the ground truth query. This metric allows for different query formulations that yield the same result. It is calculated as:
\[
\text{EX} = \frac{|\{ n \in N \mid \text{E}(Q_{gen}) = \text{E}(Q_{\text{gt}}) \}|}{N} \times 100\%
\]
where \( N \) denotes the number of questions. \( Q_{gen} \) represents the SQL query generated by the model, while \( Q_{\text{gt}} \) is the ground truth answer. \( \text{E}\) is the execution function. 

\noindent\textbf{Execution Proportion (EP)} is an auxiliary metric we proposed, evaluating the proportion of generated SQL queries that can be executed on the corresponding database without syntax errors. This metric reflects the model's upper-bound capability by assuming that any executable query is potentially correct. It is defined as:
\[
\text{EP} = \frac{|\{ n \in N \mid \text{E}(Q_{\text{gen}}) \neq \text{error} \}|}{N} \times 100\%
\]

\subsubsection{Baselines}

\noindent\textbf{Direct Response (DR)} directly generates an SQL query from the natural language question without applying any refinement techniques. The process follows a single-turn interaction.

\noindent\textbf{First Executable Query (FEQ)} leverages the model’s ability to generate multiple SQL candidates. Among candidates, the first executable query is selected without any refinement. This approach simulates multi-turn query generation.

\noindent\textbf{MAC-SQL} \citep{macsql} is an LLM-based multi-stage framework, featuring a core Decomposer agent for SQL generation supported by auxiliary agents for sub-database acquisition and query refinement. It also utilizes few-shot chain-of-thought reasoning to enhance generation processes.

\noindent\textbf{CHESS} \citep{chess} comprises four specialized agents: Information Retriever, Schema Selector, Candidate Generator, and Unit Tester. Notably, it employs locality-sensitive hashing and vector databases to efficiently retrieve relevant data from extensive database values and catalogs.

\subsubsection{Implementation Details}
All experiments ran on 4×NVIDIA A6000 with vLLM; for multi-candidate/selection we used T=0.2, top-p=0.8, while schema pruning and schema linking adopted greedy decoding for determinism

\noindent \textbf{Backbone Models.}  
Our implementation leverages both general-purpose chat models and SQL fine-tuned models. The chat models include Qwen2.5-0.5B, Qwen2.5-1.5B, Qwen2.5-Coder-1.5B \citep{qwen2.5coder}, Yi-Coder-1.5B \citep{yi-coder}, DeepSeek-Coder-1.5B \citep{deepseek}, Phi3-mini-3.8B \citep{phi3}, and MiniCPM3-4B \citep{minicpm}, while the SQL-tuned models consist of Prem-SQL-1.3B \citep{premsql} and CodeS-3B \citep{codes}.

\noindent \textbf{Candidate Size.}  
In the multi-candidate generation stage, we generate 4 candidates per path, resulting in a total candidate pool of 16. During the correction stage, the candidate size is reduced to 2.

\noindent \textbf{Selection Rounds.}  
During the selection stage, we perform 3 rounds for each selection. The final choice is the majority vote across the three rounds, ensuring consistency of the selected candidate.

\subsection{Main Results}

\subsubsection{Feather-SQL}
To validate the general effectiveness of Feather-SQL for SLMs, we conducted experiments on two datasets across a range of models (all results here were obtained using a unified model without adopting the collaboration paradigm). 

\noindent\textbf{BIRD Results.} 
As shown in Table \ref{tab:bird}, Feather-SQL demonstrates superior performance across all general-purpose chat models, achieving the highest scores in both EX and EP. For SQL fine-tuned models, Feather-SQL combined with CodeS achieves substantial gains in both EX and EP, while Prem-SQL shows notable improvements specifically in EP.

Moreover, we observe that CHESS and MAC-SQL do not perform effectively on SLMs, with their results on Qwen2.5 and Yi-Coder showing even lower EX and EP scores compared to DR. Their performance also falls behind that of FEQ.
\begin{table*}[htbp]
    \centering
    \tabcolsep=3mm
    \renewcommand{\arraystretch}{1}
    \scalebox{0.9}{
        \begin{tabular}{lcc|cc|cc}
            \toprule
            \multirow{2}{*}{\textbf{Method}} 
            & \multicolumn{2}{c}{\textbf{Qwen2.5-0.5B}} 
            & \multicolumn{2}{c}{\textbf{Yi-Coder-1.5B}} 
            & \multicolumn{2}{c}{\textbf{DeepSeek-Coder-1.3B}} \\
            & \textbf{EX (\%)} & \textbf{EP (\%)} 
            & \textbf{EX (\%)} & \textbf{EP (\%)} 
            & \textbf{EX (\%)} & \textbf{EP (\%)} \\
            \midrule
            \textbf{DR}      
                & 6.71 & \underline{26.99} 
                & 15.84 & 54.82 
                & 29.27 & 64.41\\
            \textbf{FEQ}     
                & \underline{9.65} & 29.14 
                & \underline{18.71} & \underline{73.60} 
                & \underline{30.38} & 67.67\\
            \textbf{MAC-SQL} 
                & 2.54 & 26.40 
                & 7.63  & 59.52 
                & 29.99 & \underline{77.64}\\
            \textbf{CHESS}   
                & 0.91 & 4.82 
                & 2.48  & 7.82  
                & 18.12 & 32.97\\
            \rowcolor{graybg}\textbf{Feather-SQL (Ours)} 
                & \textbf{12.52} & \textbf{30.46} 
                & \textbf{25.23} & \textbf{90.61} 
                & \textbf{36.19} & \textbf{83.70}\\
            \midrule
            \multirow{2}{*}{\textbf{Method}} 
            & \multicolumn{2}{c}{\textbf{MiniCPM3-4B}} 
            & \multicolumn{2}{c}{\textbf{Prem-SQL-1.3B}} 
            & \multicolumn{2}{c}{\textbf{CodeS-3B}} \\
            & \textbf{EX (\%)} & \textbf{EP (\%)} 
            & \textbf{EX (\%)} & \textbf{EP (\%)} 
            & \textbf{EX (\%)} & \textbf{EP (\%)} \\
            \midrule
            \textbf{DR}     
                & 27.57 & 69.30  
                & 47.07 & 88.14 
                & 24.19 & \underline{59.32}\\
            \textbf{FEQ}    
                & 29.34 & 63.89  
                & \textbf{51.63} & \underline{92.70} 
                & 25.03 & 57.50\\
            \textbf{MAC-SQL} 
                & \underline{37.35} & \underline{81.68} 
                & 8.67 (\textit{8.87\textsuperscript{\textbf{*}}})  
                & 17.01 (\textit{19.23\textsuperscript{\textbf{*}}})  
                & 10.10 (\textit{13.23\textsuperscript{\textbf{*}}})  
                & 40.87 (\textit{56.26\textsuperscript{\textbf{*}}})\\
            \textbf{CHESS}   
                & 28.42 & 54.43  
                & 24.64 & 43.22 
                & \underline{26.53} & 56.91\\
            \rowcolor{graybg}\textbf{Feather-SQL (Ours)} 
                & \textbf{40.09} & \textbf{87.02} 
                & \underline{49.28} & \textbf{98.04} 
                & \textbf{33.96} & \textbf{85.31}\\
            \bottomrule
        \end{tabular}
    }
    \caption{Comparison of EX (Execution Accuracy) and EP (Execution Proportion) across different methods on the BIRD DEV dataset. The best and second-best results are highlighted by \textbf{Bold} and \underline{underline}, respectively. \textbf{$^*$} denotes results with the extraction strategy.}
    \label{tab:bird}
\end{table*}
\noindent\textbf{Spider Results.}  Table~\ref{tab:spider} highlights the results on the Spider TEST split. Although MAC-SQL and CHESS show inconsistent performance across models, MAC-SQL generally performs well. Notably, for SQL fine-tuned models, MAC-SQL could achieve the best EX if extraction is applied, highlighting the necessity of this step. This may be attributed to MAC-SQL’s Selector mechanism, which also employs schema pruning. Unlike our table pruning approach, MAC-SQL adopts column pruning, which may be more effective for SPIDER’s relatively simple schema structures.

\begin{table*}[htbp]
    \vspace{-1cm}
    \centering
    \tabcolsep=3mm
    \renewcommand{\arraystretch}{1}
    \scalebox{0.9}{
\begin{tabular}{lcc|cc|cc}
    \toprule
    \multirow{2}{*}{\textbf{Method}}
    & \multicolumn{2}{c|}{\textbf{Qwen2.5-0.5B}} 
    & \multicolumn{2}{c|}{\textbf{Yi-Coder-1.5B}} 
    & \multicolumn{2}{c}{\textbf{DeepSeek-Coder-1.3B}} \\
    & \textbf{EX (\%)} & \textbf{EP (\%)} 
    & \textbf{EX (\%)} & \textbf{EP (\%)} 
    & \textbf{EX (\%)} & \textbf{EP (\%)} \\
    \midrule
    \textbf{DR}         & 28.50 & 56.45 & 45.23 & \underline{87.24} &   49.28  &  90.68 \\
    \textbf{FEQ}        & \underline{36.53} & 67.35 & \underline{48.30} & 86.77 &   45.46   &  89.89 \\
    \textbf{MAC-SQL}    & 29.06 & \textbf{89.61} & 13.04 & 21.70 & \textbf{52.12} & \underline{93.62} \\
    \textbf{CHESS}      & 15.42 & 29.16 & 3.68  & 10.29 & 30.18 & 46.30 \\
    \rowcolor{graybg}\textbf{Feather-SQL (Ours)} 
                        & \textbf{36.98} & \underline{75.08} & \textbf{49.56} & \textbf{92.04} & \underline{51.19} & 
                        \textbf{94.13} \\
    \midrule
    \multirow{2}{*}{\textbf{Method}} 
    & \multicolumn{2}{c|}{\textbf{MiniCPM3-4B}} 
    & \multicolumn{2}{c|}{\textbf{Prem-SQL-1.3B}} 
    & \multicolumn{2}{c}{\textbf{CodeS-3B}} \\
    & \textbf{EX (\%)} & \textbf{EP (\%)} 
    & \textbf{EX (\%)} & \textbf{EP (\%)} 
    & \textbf{EX (\%)} & \textbf{EP (\%)} \\
    \midrule
    \textbf{DR}         & 55.10 & \underline{93.71} & 60.92 & 85.79 & 47.74 & 64.23 \\
    \textbf{FEQ}        & \underline{55.75} & 89.52 & \underline{64.23} & 85.75 & 49.60 & 64.65 \\
    \textbf{MAC-SQL}    & 25.01 & 38.47 & 0.14 (\textit{{67.91}\textsuperscript{\textbf{*}}})  
                        & 0.14 (\textit{{100}\textsuperscript{\textbf{*}}})  
                        & 0 (\textit{{74.48}\textsuperscript{\textbf{*}}})  
                        & 0 (\textit{{100}\textsuperscript{\textbf{*}}}) \\
    \textbf{CHESS}      & 56.73 & 89.99 & 63.86 & \underline{92.08} & \textbf{66.65} & \underline{88.54} \\
    \rowcolor{graybg}\textbf{Feather-SQL (Ours)} 
                        & \textbf{58.92} & \textbf{94.18} & \textbf{66.60} & \textbf{92.78} & \underline{63.25} & \textbf{88.96}\\
    \bottomrule
\end{tabular}
}

    \caption{Comparison of EX (Execution Accuracy) and EP (Execution Proportion) across different methods on the Spider TEST dataset. The best and second-best results for EX are highlighted by \textbf{bold} and \underline{underline}, respectively. \textbf{$^*$} denotes results with the extraction strategy.}
    \vspace{-3mm}
    \label{tab:spider}
\end{table*}

\subsubsection{1+1 Collaboration Paradigm}

As observed in Table \ref{tab:bird}, although Feather-SQL improves the EP of Prem-SQL, its EX shows a 2\% decrease compared to FEQ. This decline is primarily due to Prem-SQL's inability to handle auxiliary reasoning tasks. To address this limitation, we propose a division of tasks where the general-purpose chat model handles auxiliary reasoning, while the SQL fine-tuned model focuses on SQL generation.

As shown in Table~\ref{tab:paradigm-feather}, our 1+1 collaboration paradigm under Feather-SQL achieves a 3--6\% improvement in EX for both Prem-SQL and CodeS. However, we observe a decline in EP when paired with a chat model. This is because when the SQL model is also used as the chat model during schema pruning, it returns a query instead of the expected answer. But our extraction strategy sitll retrieves table names from the output, often resulting in an overly pruned schema-containing only one or two tables. While a simplified schema can occasionally boost EP, it frequently leads to lower overall EX.

Additionally, Table~\ref{tab:paradigm-chess} shows that our paradigm improves both Prem-SQL and CodeS in CHESS, with EX increasing by \textasciitilde20\% and EP by over \textasciitilde35\% for Prem-SQL, while CodeS sees a smaller but consistent EX gain with no clear trend in EP.

\begin{table*}[htbp]
   \vspace{-2mm}
  \centering

  \begin{subtable}[t]{0.48\textwidth}

    \centering
    \renewcommand{\arraystretch}{1.1}
    \resizebox{\linewidth}{!}{
      \begin{tabular}{ll|cc}
        \toprule
        \textbf{Chat Model} & \textbf{SQL Model} & \textbf{EX (\%)} & \textbf{EP (\%)} \\
        \midrule
        -- & Prem-SQL & 49.28 & 98.04 \\
        Qwen & Prem-SQL & 52.44 $\uparrow$ & 94.08 \\
        Qwen Coder & Prem-SQL & 52.83 $\uparrow$ & 98.31 \\
        Yi Coder & Prem-SQL & 54.76 $\uparrow$ & 93.94 \\ 
        \midrule
        -- & CodeS & 33.96 & 83.31 \\ 
        Qwen & CodeS & 35.79 $\uparrow$ & 80.05 \\ 
        Qwen Coder & CodeS & 37.03 $\uparrow$ & 81.10 \\
        Yi Coder & CodeS & 39.43 $\uparrow$ & 80.44 \\
        \bottomrule
      \end{tabular}
    }
    \caption{Feather-SQL}
    \label{tab:paradigm-feather}
  \end{subtable}\hfill
  \begin{subtable}[t]{0.48\textwidth}
    \centering
    \renewcommand{\arraystretch}{1.1}
    \resizebox{\linewidth}{!}{
      \begin{tabular}{ll|cc}
        \toprule
        \textbf{Chat Model} & \textbf{SQL Model} & \textbf{EX (\%)} & \textbf{EP (\%)} \\
        \midrule
        -- & Prem-SQL & 24.64 & 43.22 \\
        Qwen & Prem-SQL & 49.28 $\uparrow$ & 82.07 \\
        Qwen Coder & Prem-SQL & 49.61 $\uparrow$ & 79.60 \\
        Yi Coder & Prem-SQL & 47.65 $\uparrow$ & 79.79 \\
        \midrule
        -- & CodeS & 26.53 & 56.91 \\ 
        Qwen & CodeS & 28.55 $\uparrow$ & 56.19 \\
        Qwen Coder & CodeS & 28.88 $\uparrow$ & 63.04 \\
        Yi Coder & CodeS & 27.44 $\uparrow$ & 55.22 \\
        \bottomrule
      \end{tabular}
    }
    \caption{CHESS}
    \label{tab:paradigm-chess}
  \end{subtable}
    \caption{Paradigm performance on the BIRD DEV dataset. When no chat model is specified, the SQL model is also used as the chat model. Qwen refers to Qwen2.5-1.5B, Qwen Coder refers to Qwen2.5-Coder-1.5B, Yi Coder refers to Yi-Coder-1.5B, Prem-SQL refers to Prem-SQL-1.3B, and CodeS refers to CodeS-3B.}
    \vspace{-6mm}    
\end{table*}

However, the two models benefit differently due to their handling of auxiliary tasks. Prem-SQL attempts to answer linking questions but often does so incorrectly, whereas CodeS, due to severe catastrophic forgetting, fails to provide valid responses. As a result, CHESS defaults to using the original schema with CodeS, reducing linking errors.

Furthermore, since CHESS constructs long prompts without schema pruning, introducing a chat model increases input length and complexity. While this improves reasoning, it does not fully offset CodeS’s limitations in processing extended inputs, restricting its EX improvement.

\begin{figure}[t]
  \vspace{-1.5cm}
  \centering
  \includegraphics[width=\linewidth]{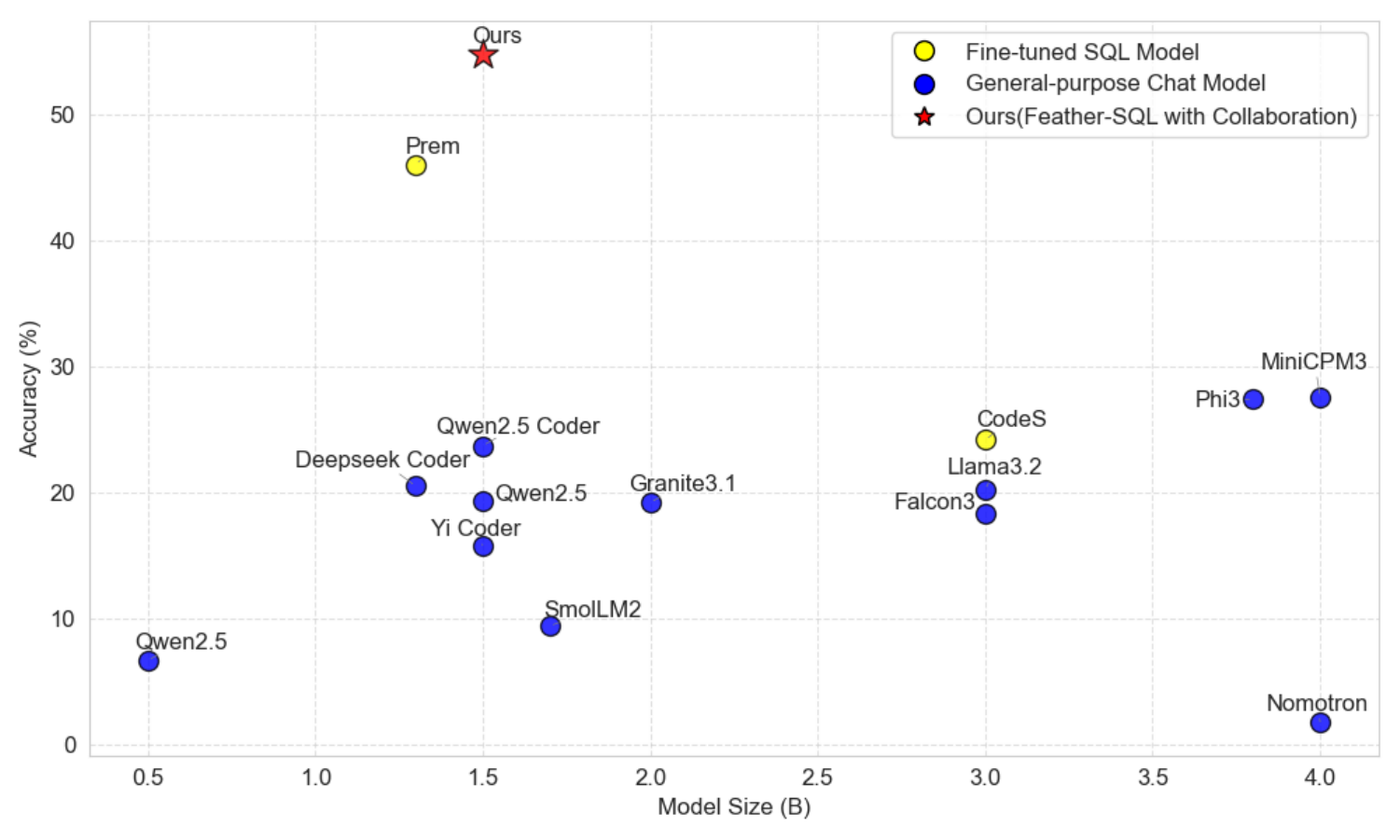}
  \caption{Accuracy (\%) versus model size (billions of parameters) on BIRD DEV for small language models.
  Fine-tuned SQL models are shown in yellow, general-purpose chat models in blue, and ours (Feather-SQL + 1+1 Model Collaboration) is marked with a red star.}
  \label{fig:sota}
  \vspace{-0.6cm}
\end{figure}

\noindent\textbf{SOTA within SLMs.}
To contextualize our results within the small-model landscape, we further compare against widely used open-source SLMs beyond our backbones—namely Granite-3.1B~\citep{granite}, SmolLM2-1.7B, Llama3.2-3B~\citep{llama3}, Falcon-3B~\citep{falcon}, and Nemotron-4B~\citep{nemotron}. Figure~\ref{fig:sota} shows that combining Feather-SQL with the 1+1 Model Collaboration paradigm yields state-of-the-art accuracy among small language models.

\begin{wraptable}{r}{0.5\textwidth}
\vspace{-2.1cm}
\centering
\renewcommand{\arraystretch}{1.2}
\tabcolsep=2mm
\begin{tabular}{l|cc}
\toprule
\textbf{Framework} & \textbf{EX (\%)} & \textbf{EP (\%)} \\ 
\midrule
Full Model & 31.81 & 88.33 \\ 
--w/o Schema Pruning & -4.63 $\downarrow$ & -20.34 $\downarrow$ \\ 
--w/o Schema Linking & -3.45 $\downarrow$ & -20.92 $\downarrow$ \\ 
--w/o Multi-Candidate & -2.47 $\downarrow$ & -17.99 $\downarrow$ \\ 
--w/o Correction & -0.20 $\downarrow$ & -12.58 $\downarrow$ \\ 
--w/o Selection & -2.21 $\downarrow$ & -10.36 $\downarrow$ \\ 
\bottomrule
\end{tabular}
\caption{Ablation Study on Framework Components.}
\label{tab:component}
\vspace{-0.6cm}
\end{wraptable}

\vspace{-2mm}
\subsection{Ablation Studies}

\subsubsection{Component Contribution}
\label{sec:cc}

We conducted an ablation study to quantify the impact of each framework component by removing them one at a time and measuring changes in EX and EP on the BIRD DEV dataset, using Qwen2.5-1.5B (Table~\ref{tab:component}). 

We can see from the ablation results that removing any of the components causes a drop in both EX and EP. This underscores that each step in our pipeline contributes to overall performance, and omitting even one module leads to noticeably reduced accuracy or executability.

Among these, schema pruning is shown to be the most critical: when it is removed, EX falls from 31.81\% to 27.18\%, the single largest drop in our study. This highlights how focusing on only the relevant tables and columns helps the model concentrate on essential schema elements, thereby yielding more accurate SQL generation. In contrast, removing correction only reduces EX by 0.20\%, indicating that it has a relatively minor impact on the framework’s effectiveness.

\subsubsection{Path Contribution}

We analyzed the origins of SQL answers from four models to understand how each processing path affects the final output.

\newpage
\vspace{-0.8cm}
\begin{wrapfigure}{r}{0.5\linewidth}
\vspace{-12mm}
\includegraphics[width=\linewidth]{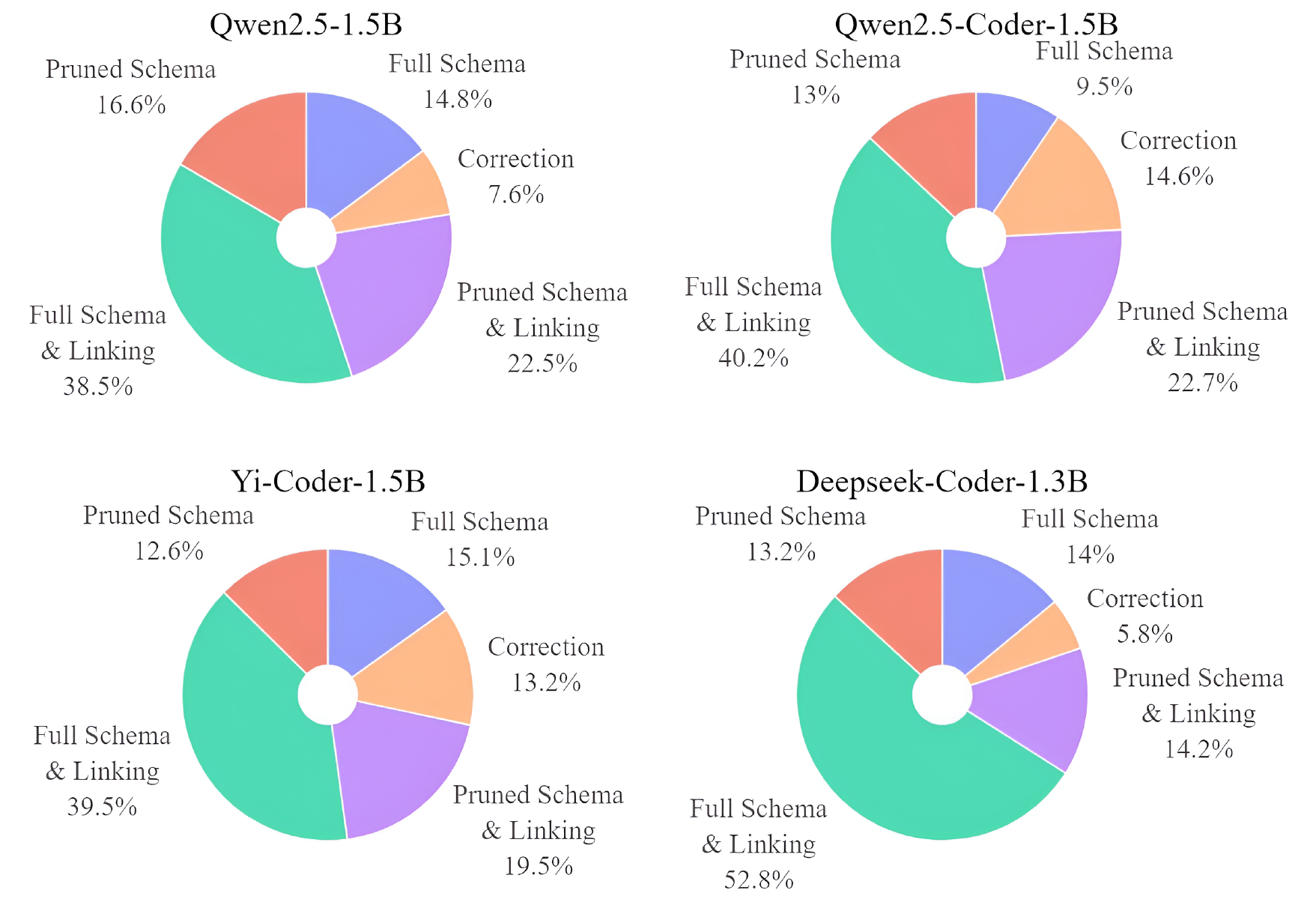}
\caption{Distribution of correct SQL answers contributed by each path across four different SLMs.}
\vspace{-10mm}
\label{fig:distribution}
\end{wrapfigure}

As shown in Figure \ref{fig:distribution}, our multi-path framework includes four paths:
one using both schema linking and pruning, one using only schema linking, one using only schema pruning, and one without either.

For all four models, the path \textit{Full Schema \& Linking} is consistently the largest contributor, followed by \textit{Pruned Schema \& Linking}. This ranking underscores the critical role of linking in the framework, regardless of whether the schema is pruned or not.

Additionally, we find that schema pruning collectively accounts for over 25\% across the models. These observations are consistent with the ablation findings in \ref{sec:cc}, further illustrating the essential roles of each component in ensuring executable and accurate query generation.

\subsubsection{Candidate Size}

\begin{wrapfigure}{r}{0.5\linewidth}
\vspace{-6mm}
\includegraphics[width=1.1\linewidth]{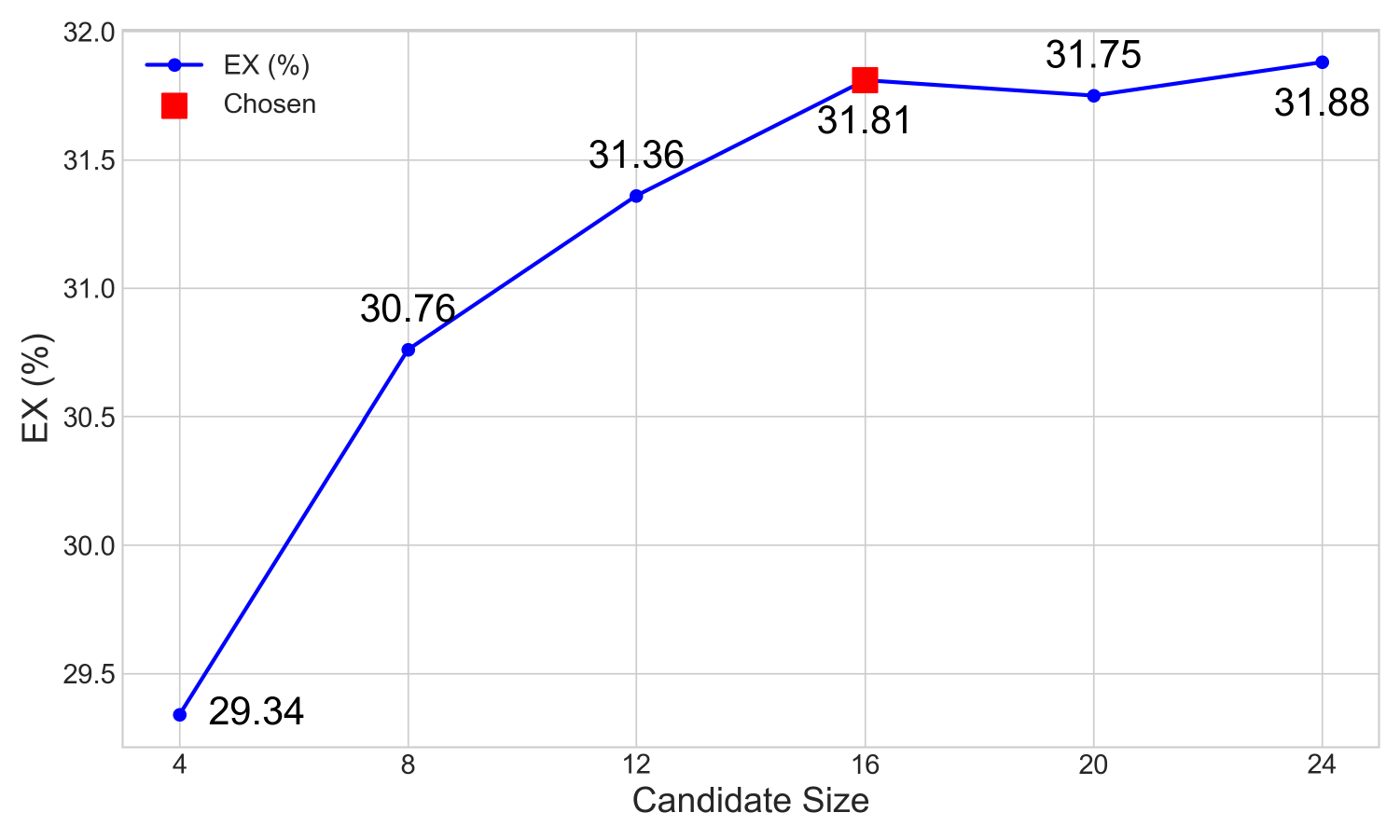}
\caption{Effect of candidate size on EX performance. }
\label{fig:candidate}
\vspace{-3mm}
\end{wrapfigure}

We further investigated the impact of different candidate sizes. Figure~\ref{fig:candidate} presents the results based on our four paths. In our experiments, the total candidate size increases from 4 to 24, which corresponds to the number of candidates generated per path increasing from 1 to 6. The figure illustrates how EX changes as the overall candidate size grows from 4 to 24.

We observe a concave trend, consistent with Figure~\ref{fig:gains}: EX steadily increases as the candidate size rises from 4 to 16 but then plateaus from 16 to 24. Once the model reaches its approximate upper bound, further increases in candidate size result in only a marginal difference in performance. Therefore, we select a candidate size of 16, as it is the earliest point at which EX saturates, thus balancing computational efficiency and model performance.

\begin{wraptable}{r}{0.3\linewidth}
\vspace{-18mm}
\centering
\renewcommand{\arraystretch}{1.15}
\begin{tabular}{lc}
\toprule
\textbf{Rounds $R$} & \textbf{EX (\%)} \\
\midrule
1 & 29.40 \\
\textbf{3} & \textbf{31.81} \\
5 & 31.23 \\
7 & 31.49 \\
\bottomrule
\end{tabular}
\caption{Effect of selection rounds on EX performance.}
\label{tab:sel_rounds}
\vspace{-1.5cm}
\end{wraptable}

\subsubsection{Selection Rounds}

We repeat the selection step multiple times and choose the SQL that appears the most frequently (mode). As shown in Table~\ref{tab:sel_rounds}, EX improves from a single round to three rounds, after which it largely plateaus. Accordingly, we use three rounds by default.

\section{Limitations}
\label{sec:limitations}

\subsection{Framework Upper Bound}
\begin{wraptable}{r}{0.5\linewidth}
\vspace{-12mm}
\centering
\renewcommand{\arraystretch}{1.2}
\resizebox{\linewidth}{!}{
\begin{tabular}{lccc}
\toprule
\textbf{Model} & \textbf{Top-1 (\%)} & \textbf{Top-2 (\%)} & \textbf{Top-3 (\%)} \\ 
\midrule
Qwen       & 31.8 & 39.0 & 40.5 \\ 
Yi Coder   & 25.2 & 32.6 & 34.5 \\ 
Prem-SQL   & 49.2 & 60.2 & 62.6 \\ 
\bottomrule
\end{tabular}
}
\caption{Cumulative accuracy on BIRD DEV. Top-$N$ is the percentage of questions with the correct SQL among the $N$ best candidates. Qwen refers to Qwen2.5-1.5B.}
\label{tab:cumulative_accuracy_comparison}
\vspace{-2mm}
\end{wraptable}
Our current selection mechanism in Feather-SQL is limited in fully exploiting the quality of generated candidates.
As shown by Table~\ref{tab:cumulative_accuracy_comparison}, Top-3 is roughly 10\% higher than Top-1.
This gap indicates that, while the framework can produce correct SQL among its outputs, the selection stage does not always identify the optimal one. More accurate or adaptive selection strategies could narrow this gap, thereby raising overall performance.

\subsection{Remaining Accuracy Gap}
\label{sec:accuracy-ceiling}
Feather-SQL delivers clear gains at the 1B-parameter scale, yet a persistent gap remains between small language models (SLMs) and large language models (LLMs). On BIRD, leading LLM-based NL2SQL approaches report around 75\% accuracy~\citep{xiyan,chase-sql} and many LLM systems typically achieve 60\%+, whereas state-of-the-art performance of SLMs remain comparatively lower. Closing this gap will require further advances in reasoning, schema understanding, and selection mechanisms tailored to SLMs.

\section{Conclusion}
In this work, we introduced Feather-SQL, the first lightweight framework designed to enhance NL2SQL performance for SLMs. We conduct comprehensive evaluations on the challenging BIRD and Spider datasets, where Feather-SQL yields improvements in both executability and accuracy. Additionally, we present the 1+1 Model Collaboration paradigm—a novel approach that pairs a general-purpose chat model with a SQL specialist to combine robust reasoning with precise query generation. Our evaluation results show that this paradigm boosts accuracy across different frameworks, demonstrating its consistent effectiveness. Moreover, the flexibility of our approach provides a robust foundation not only for advancing NL2SQL but also for application to other structured tasks and domains. Together, our work outline a practical path toward trustworthy, lightweight, edge-deployable database querying.

\newpage
\onecolumn
\bibliography{tmlr}
\bibliographystyle{tmlr}

\newpage
\appendix

\section{Multi-Candidate Motivation}
\label{app:mutli-candidate}
\begin{table*}[htbp]
    \centering
    \tabcolsep=3mm
    \renewcommand{\arraystretch}{1}
    \scalebox{0.95}{
    \begin{tabular}{lcc|cc|cc}
        \toprule
        \multirow{2}{*}{\textbf{Top-N}} & \multicolumn{2}{c}{\textbf{Yi-Coder-1.5B}} & \multicolumn{2}{c}{\textbf{MiniCPM3-4B}} & \multicolumn{2}{c}{\textbf{Prem-SQL-1.3B}}\\
         & \textbf{ACC (\%)} & \textbf{EXE (\%)} & \textbf{ACC (\%)} & \textbf{EXE (\%)} & \textbf{ACC (\%)} & \textbf{EXE (\%)}\\
        \midrule
        1           & 15.65   & 46.26   & 26.53   & 65.31   & 55.78   & 92.52   \\
        3           & 24.49   & 70.75   & 35.37   & 76.87   & 59.86   & 97.28   \\
        5           & 30.61   & 78.91   & 36.05   & 82.31   & 62.59   & 97.96   \\
        7           & 33.33   & 82.31   & 37.41   & 84.35   & 65.31   & 97.96   \\
        \midrule
        \multirow{2}{*}{\textbf{Top-N}} & \multicolumn{2}{c}{\textbf{CodeS-3B}} & \multicolumn{2}{c}{\textbf{GPT-4o}} & \multicolumn{2}{c}{\textbf{Claude-3.5-Sonnet}}\\
         & \textbf{ACC (\%)} & \textbf{EXE (\%)} & \textbf{ACC (\%)} & \textbf{EXE (\%)} & \textbf{ACC (\%)} & \textbf{EXE (\%)}\\
        \midrule
        1           & 24.49   & 61.90   & 51.70   & 93.20   & 40.82   & 86.39   \\
        3           & 27.21   & 68.71   & 53.74   & 94.56   & 41.50   & 87.76   \\
        5           & 29.93   & 72.11   & 56.46   & 94.56   & 42.18   & 88.44   \\
        7           & 29.93   & 73.47   & 56.46   & 94.56   & 42.18   & 88.44   \\
        \bottomrule
    \end{tabular}
    }
    \caption{Comparison of Accuracy (ACC) and Execution (EXE) on the BIRD DEV Subset from CHESS using multi-candidate generation strategy. }
    \label{tab:multi-candidate-results}
\end{table*}

\section{Prompts}
\subsection{}
\begin{tcolorbox}[
  title=Schema Pruning Prompt,
  colback=blue!5!white,
  colframe=blue!75!black,
  breakable,
  enhanced jigsaw,   
  sharp corners,     
  boxrule=1pt        
]
\lstset{
  basicstyle=\ttfamily\footnotesize, 
  breaklines=true,
  columns=fullflexible,   
  keepspaces=true
}

\begin{lstlisting}
prompt_pruning_system = """
You are an agent designed to find all related tables to generate SQL query 
for question based on the database schema and hint.

## Requirements
1. You don't need to answer the question, your task is only finding all related tables .
2. Consider all constraints of each table, including primary keys, foreign keys, and data types.
3. You can generate chain of thoughts, but ensure all tables mentioned truly exist.
4. Successfully answer related columns could help you win $100000 dollars.
"""

prompt_pruning = """
## Instructions
1. Prioritize the table that most directly contains the information needed to answer the question, considering:
    - Table relationships such as foreign keys.
    - Whether the table has columns directly related to the entities or actions in the question.
2. Reasoning like two shown examples.

----------Example----------
## Database Schema
CREATE TABLE Employees (
    employee_id INT PRIMARY KEY,
    name VARCHAR(100),
    department VARCHAR(100),
    salary DECIMAL(10, 2)
);

CREATE TABLE Departments (
    department_id INT PRIMARY KEY,
    department_name VARCHAR(100),
    location VARCHAR(100)
);

## Question
What is the salary of the employee named 'Alice'?

## Relevant Tables
This table directly contains the columns name and salary, which are the only necessary fields to answer the question.
The name column is used to locate the specific employee named 'Alice', and the salary column provides the required 
salary information. The Departments table is irrelevant because it does not store employee-level data like salaries 
or names, and its information is unrelated to this specific query.
The relevant table is Employees.

----------Task----------
## Database Schema
You are provided with the structure of the database "{database_name}":
{database_schema}

## Question
{question}

## Hint
{hint}

Among the following tables: {tables}, which tables are relevant for addressing the question?
## Relevant Tables
"""
\end{lstlisting}
\end{tcolorbox}

\subsection{}
\begin{tcolorbox}[
  title=Schema Linking Prompt,
  colback=blue!5!white,
  colframe=blue!75!black,
  breakable,
  enhanced jigsaw,   
  sharp corners,     
  boxrule=1pt        
]
\lstset{
  basicstyle=\ttfamily\footnotesize, 
  breaklines=true,
  columns=fullflexible,   
  keepspaces=true
}

\begin{lstlisting}
prompt_linking_system="""
You are an agent designed to find all related columns to generate SQL query for question based on the database schema and the hint.

## Requirements
1. You don't need to answer the question, your task is only finding all related columns.
2. Hint could help you to find the correct related columns.
3. Consider all constraints of each table, including primary keys, foreign keys, and data types.
4. You can generate chain of thoughts, but ensure all columns mentioned truly exist.
7. Successfully answer related columns could help you win $100000 dollars.
"""

prompt_linking="""
## Instructions
1. Select columns that relates to information requested by the question, considering:
    - Whether the column is key to filtering results (used in WHERE clauses).
    - Whether the column should be part of the SELECT statement to fulfill the user query.
    - The relationship of the column to other parts of the question, such as groupings, aggregations, or direct match to entities mentioned.
2. Reasoning like two shown examples.

----------Example----------
## Database Schema
CREATE TABLE Employees (
    employee_id INT PRIMARY KEY,
    name VARCHAR(100),
    department VARCHAR(100),
    salary DECIMAL(10, 2)
);

CREATE TABLE Departments (
    department_id INT PRIMARY KEY,
    department_name VARCHAR(100),
    location VARCHAR(100)
);

## Question
What is the salary of the employee named 'Alice'?

## Relevant Columns
The name column is essential to filter the employee named 'Alice' in the WHERE clause, ensuring we identify the correct individual. The salary column is needed to extract the requested information, which is the employee's salary. Since the question does not involve departments, the Departments table and its columns are irrelevant.
The related columns are Employees.name and Employees.salary.

----------Task----------
## Database Schema
You are provided with the structure of the database "{database_name}":
{schema}

## Question
{question}

## Hint
{hint}

Among the columns, which are relevant for addressing the question?
## Relevant Columns
"""
\end{lstlisting}
\end{tcolorbox}

\subsection{}
\begin{tcolorbox}[
  title=Multi-path Generation Prompt,
  colback=blue!5!white,
  colframe=blue!75!black,
  breakable,
  enhanced jigsaw,   
  sharp corners,     
  boxrule=1pt        
]
\lstset{
  basicstyle=\ttfamily\footnotesize, 
  breaklines=true,
  columns=fullflexible,   
  keepspaces=true
}

\begin{lstlisting}
system_prompt_sql_generation = """
You are an expert SQL assistant tasked with generating precise SQL queries based on given database schemas, questions, and hint.

## Responsibilities
1. Analyze the **database schema** and **hint** to determine relationships, including **primary keys, foreign keys, data types, and constraints**.
2. Generate a single, valid **SQLite SQL query** to answer the question, using provided schema linking information for table and column selection.
3. Your response should contain only the **SQL query**, using standard SQL syntax with correct use of table/column names and SQL clauses.

## Requirements
- Respond with only one SQL query, formatted as ```SQL```.
- Use clauses like **SELECT**, **FROM**, **WHERE**, **JOIN**, **GROUP BY**, **ORDER BY**, etc.
- Ensure SQL is efficient and respects **Important Columns**, table relationships, and relevant constraints.
"""

prompt_generation_with_linking = """
You are given a database schema, question, important columns and hint. Generate a valid SQLite query that answers the question.

## Instructions
1. Your response should only contain one SQL query, in standard SQL syntax.
2. Consider all **table relationships**, **primary/foreign keys**, **data types**, and **Important Columns** while generating the query.

## Database Schema
Database "{database_name}":
{database_schema}

## Important Columns
{schema_linking}

## Question
{question}

## Hint
{hint}

## Output Requirement
Format the response as:
```sql
[SQL query]
```
"""

prompt_generation_without_linking = """
You are given a database schema, question, and hint. Generate a valid SQLite query that answers the question.

## Instructions
1. Your response should only contain one SQL query, in standard SQL syntax.
2. Consider all **table relationships**, **primary/foreign keys**, **data types** while generating the query.

## Database Schema
Database "{database_name}":
{database_schema}

## Question
{question}

## Hint
{hint}

## Output Requirement
Format the response as:
```sql
[SQL query]
```
"""
\end{lstlisting}
\end{tcolorbox}
\subsection{}
\begin{tcolorbox}[
  title=Correction Prompt,
  colback=blue!5!white,
  colframe=blue!75!black,
  breakable,
  enhanced jigsaw,   
  sharp corners,     
  boxrule=1pt        
]
\lstset{
  basicstyle=\ttfamily\footnotesize, 
  breaklines=true,
  columns=fullflexible,   
  keepspaces=true
}

\begin{lstlisting}
prompt_answer_correction_system ="""
Suppose you are an expert in SQLite and database management.

## Instructions
1. Based on the database structure provided, previous answer and its error messages, generate one SQL query that answers the question.
2. You should try to fix the error of the previous answer and avoid it from happening again.

## Requirements
1. Your response should consist of only one SQL query, don't generate anything else.
3. Consider all constraints of each table, including primary keys, foreign keys, and data types.
4. Provide your query in standard SQL format with appropriate use of SQL functions, joins, and conditions.
"""

prompt_answer_correction = """
## Database Schema
Given the structure of database:
{schema}

## Question
{question}

## Hint
{hint}

## Previous answer
{prev_ans}

## Error
{errorMsg}

## New Answer
""" 
\end{lstlisting}
\end{tcolorbox}

\subsection{}
\begin{tcolorbox}[
  title=Selection Prompt,
  colback=blue!5!white,
  colframe=blue!75!black,
  breakable,
  enhanced jigsaw,   
  sharp corners,     
  boxrule=1pt        
]
\lstset{
  basicstyle=\ttfamily\footnotesize, 
  breaklines=true,
  columns=fullflexible,   
  keepspaces=true
}

\begin{lstlisting}
system_prompt_query_selection = """
You are an expert in analyzing SQL queries and determining their relevance to a given question. Your task is to evaluate multiple SQL queries and select the one that best answers the question based on the provided database schema and context.

## Responsibilities
1. Analyze the given question: Understand the intent of the question and its expected output.
2. Evaluate each SQL query: Consider the correctness, relevance, and completeness of each query in relation to the question.
3. Select the best query: Choose the query that most accurately answers the question, while considering database structure, table relationships, and query efficiency.

## Requirements
- Respond with the most relevant SQL query, and nothing else.
- Ensure the selected query is valid for the given database schema and directly addresses the question.
"""

query_selection_prompt = """
You are given a question, a database schema, and multiple SQL queries. Your task is to select the SQL query that is most relevant and best answers the question.

## Instructions
1. Analyze the Question: Understand what the user is asking and identify the information that needs to be extracted from the database.
2. Evaluate SQL Queries: For each provided SQL query, determine its relevance based on:
    - Accuracy: Does the query correctly match the question's intent?
    - Completeness: Does the query retrieve all the necessary information without omitting important details?
    - Efficiency: Is the query optimized for the task, avoiding unnecessary joins or conditions?
3. Select the Most Relevant Query: Choose the query that is the best match for the question.

## Database Schema
Database "{database_name}":
{database_schema}

## Question
The question is:
{question}

## Hint
{hint}

## SQL Queries
{queries}

## Output Requirement
Reply the query Index in the format of "Index: ". 

## Output 
"""

query_with_response_selection_prompt = """
You are given a question, a database schema, multiple SQL queries, and their execution results. Your task is to select the SQL query that best answers the question based on the query and its result.

## Instructions
1. Understand the Question: Determine what the user is asking and identify the specific information that needs to be retrieved.
2. Evaluate Each Query and Response Pair: For each provided SQL query and its result, determine:
    - Query Accuracy: Does the query correctly represent the user's intent?
    - Result Relevance: Does the result contain the data needed to answer the question completely and correctly?
    - Efficiency: Is the query optimized, avoiding unnecessary complexity?

## Database Schema
Database "{database_name}":
{database_schema}

## Question
{question}

## Hint
{hint}

## SQL Queries and Execution Results
{queries}

## Output Requirement
Only reply the query Index in the format of "Index: ". 
"""
\end{lstlisting}
\end{tcolorbox}

\end{document}